\documentclass[runningheads,a4paper]{llncs}

\usepackage[year=2026]{eccv}

\usepackage{eccvabbrv}

\usepackage{graphicx}
\usepackage{booktabs}
\usepackage{array}
\usepackage{amssymb}
\usepackage{tikz}
\usetikzlibrary{positioning, arrows.meta}

\usepackage[accsupp]{axessibility}  

\usepackage{hyperref}

\usepackage{orcidlink}

\newcommand\blfootnote[1]{{\renewcommand\thefootnote{}\footnotetext{#1}}}

\begin{document}

\title{Sim-to-Real Traffic Scene Understanding by Decoupling Semantics from Caption Generation with V-JEPA}
\titlerunning{Sim-to-Real Traffic Scene Understanding with V-JEPA}

\author{Nguyen Hoai Thuong Bui\thanks{These authors contributed equally.} \and
Thanh Nguyen Vo$^{\star}$ \and
Trinh Tra Giang Nguyen$^{\star}$ \and
Ha Duc Bui\thanks{Corresponding author.}}

\authorrunning{N.~H.~T.~Bui et al.}

\institute{Ho Chi Minh City University of Technology and Engineering (HCMUTE),\\
Ho Chi Minh City, Vietnam\\
\email{thuongbui7198@gmail.com, nguyenvothanh04@gmail.com, nguyentrinhtragiang03@gmail.com, ducbh@hcmute.edu.vn}}

\maketitle
\blfootnote{Code: \url{https://github.com/ThuongBuiRVC/Traffic-JEPA}}

\begin{abstract}
Track 2 of the AI City Challenge 2026 requires both visual question answering
(VQA) and traffic event description generation under a challenging
synthetic-to-real domain shift. Existing vision-language approaches often
entangle semantic understanding with language generation, making them
susceptible to hallucination and inconsistent reasoning across event phases. In
this work, we propose a decoupled semantic understanding framework that first
resolves predefined traffic questions into structured semantic facts and
subsequently uses these facts to guide caption generation. A frozen V-JEPA
encoder extracts predictive scene representations, while a lightweight
Llama-based predictor produces answers for VQA queries. To improve reliability,
we introduce a training-free structured refinement mechanism that exploits
statistical priors, inter-question relationships, and temporal event consistency
to correct prediction errors. The refined semantic facts are then provided to
Qwen3-VL-8B to generate pedestrian and vehicle descriptions for each traffic
event. Experimental results on the official 2026 AI City Challenge Track 2
benchmark show that the proposed method achieves 87.09\% VQA accuracy and an
overall S2 score of 60.0853, ranking first among all participating teams. These
results demonstrate that predictive world representations combined with
structured semantic refinement enable more accurate and reliable traffic
understanding, leading to higher-quality language generation.

\end{abstract}

\section{Introduction}
\label{sec:intro}

Autonomous driving requires a vehicle to continuously interpret complex traffic situations in which pedestrians, vehicles, and road infrastructure interact and evolve over time~\cite{drivelm,drivevlm}. Reliable driving decisions therefore depend not only on detecting the elements present in a scene, but also on understanding how their states, behaviors, and interactions jointly define the ongoing event. Such understanding provides the semantic context needed to recognize what is happening, assess potential hazards, and inform downstream prediction and planning.

Recent advances in vision-language models (VLMs) have significantly expanded the capabilities of traffic scene understanding by enabling open-ended question answering, semantic captioning, and language-based reasoning directly from visual observations~\cite{llava,videollama}. Compared with conventional perception pipelines that predict predefined labels, VLMs provide richer descriptions of complex traffic situations and explain the interactions among road users in natural language. Reflecting this trend, Track 2 of the 2026 AI City Challenge evaluates whether models can understand simulated driving scenes through visual question answering and comprehensive scene captioning, emphasizing both semantic accuracy and reasoning ability~\cite{Tang26AICity26,aicity2026,wts}.

Despite these advances, directly relying on VLMs to infer traffic semantics from raw video remains challenging. Driving scenes are inherently dynamic, containing multiple agents whose behaviors continuously evolve and influence one another. Understanding such scenes requires jointly reasoning about object identities, temporal evolution, spatial interactions, and causal relationships before forming a coherent semantic description. Asking a language model to perform this entire reasoning process directly from visual tokens can easily produce hallucinated events or logically inconsistent interpretations~\cite{vlmhallucination}. Moreover, Track 2 introduces an additional sim-to-real challenge~\cite{tobin2017,ganin2016} in which discrepancies in visual appearance, environmental conditions, and traffic characteristics reduce the robustness and generalizability of end-to-end language reasoning.

To address these challenges, we propose a two-stage approach in which traffic scene understanding begins with structured semantic analysis and is followed by language generation. Rather than prompting a language model to directly describe the entire traffic scene from visual observations, our approach first answers the structured visual questions provided in Track 2. Each question focuses on a specific semantic aspect of the driving scenario, such as participating agents, their attributes, behaviors, spatial relationships, or traffic events. The resulting pairs of questions and answers form a structured semantic representation that grounds the subsequent caption generation process. By reasoning over explicit semantic facts instead of raw visual tokens alone, the language model can produce more coherent and faithful descriptions while reducing hallucinations caused by complex traffic dynamics.

Specifically, we first leverage V-JEPA~\cite{vjepa21} to extract high-level scene representations that are less sensitive to appearance variations. A Llama-based predictor is then trained to infer structured semantic facts by answering the visual questions directly from these features. To eliminate contradictions, these predicted semantics are refined by a training-free mechanism that enforces temporal consistency and logical relationships among the questions. Finally, this refined semantic representation guides a Qwen model to generate the final description. This design allows language generation to serve as a means of expressing grounded scene understanding, rather than compensating for uncertain visual evidence with linguistic priors.

The main contributions of this paper are summarized as follows:
\begin{enumerate}
  \item We propose a structured traffic scene understanding framework that grounds caption generation in explicit scene semantics, which are first extracted through reasoning guided by the benchmark questions.

  \item We introduce a training-free semantic refinement mechanism that exploits relationships among questions and the temporal structure of events to improve semantic consistency and produce a coherent interpretation of the traffic event.

  \item We achieve the top-ranked performance in Track 2 on the 2026 AI City Challenge leaderboard, demonstrating the effectiveness and domain transferability of our framework under challenging sim-to-real conditions.
\end{enumerate}

\section{Related Work}
\label{sec:related-work}

\subsection{Traffic Scene Understanding and Captioning}

Recent driving VLMs have moved beyond isolated recognition outputs toward
structured language reasoning. DriveLM represents this shift through Graph VQA,
which links perception, prediction, and planning questions so that driving
decisions build on earlier scene evidence~\cite{drivelm}. DriveVLM follows the same
principle at the system level by organizing reasoning into scene description,
scene analysis, and hierarchical planning~\cite{drivevlm}. This structured view is also
important for WTS, where captions must describe fine-grained
pedestrian--vehicle interactions across multiple views and event phases~\cite{wts}.
Accordingly, Kachhadiya et al. convert question--answer outputs into explicit
facts before caption generation~\cite{kachhadiya2025}. However, facts predicted independently
can still contradict one another across related questions or event phases. Our
method addresses this limitation by refining their relational and temporal
consistency before using them, together with video frames, to guide caption
generation.

\subsection{Latent Video Representation Learning}

Self-supervised video learning seeks to capture temporal structure without
requiring task-specific annotations. While reconstruction objectives devote
capacity to recovering visual details, V-JEPA predicts the latent
representations of masked spatiotemporal regions from their visible context~\cite{vjepa,lecun2022}. V-JEPA~2 scales this principle to large video collections, producing
representations that support motion understanding and action anticipation~\cite{vjepa2}. VL-JEPA then extends latent prediction to vision-language learning by
predicting continuous text embeddings, enabling retrieval and discriminative
VQA without autoregressive decoding~\cite{vljepa}. This progression motivates our use
of frozen V-JEPA features and our formulation of answer prediction as retrieval
in a semantic embedding space.

\subsection{Parameter-Efficient Adaptation and Structured Prediction}

Adapting large pretrained models to limited task data commonly relies on
learning a small set of parameters while keeping the backbone frozen. Prompt
tuning~\cite{prompttuning} learns continuous input tokens that
condition a frozen language model, whereas LoRA introduces trainable low-rank
updates to selected model weights~\cite{lora}. Although these methods reduce training cost, they do not enforce
agreement among predictions made independently. Structured prediction provides
the complementary ability to model such dependencies through compatibility
scores and sequence decoding~\cite{kschischang2001,viterbi1967}. Our framework combines both ideas: compact
conditioning and projection modules adapt the frozen visual and language
backbones, while a training-free decoder uses statistical priors, question
relations, and phase transitions to refine the predicted answers jointly.

\section{Method}
\label{sec:method}

\subsection{Overview}
\label{sec:method-overview}

Track 2 requires both fine-grained visual question answering and the generation of pedestrian and vehicle descriptions for each traffic event. Rather than treating these as independent tasks, we exploit their complementary nature: the benchmark questions explicitly specify the semantic attributes that should appear in the final descriptions. Our framework therefore first resolves these queries into structured semantic facts and subsequently uses these facts to drive caption generation. This design is well suited to the benchmark because the questions explicitly enumerate the safety-critical attributes that should be reported, including actor awareness, gaze direction, relative position, and behavioral changes across event phases. In contrast, free-form caption generation leaves these attributes unspecified and often emphasizes visually salient but less informative content.

\begin{figure}[t]
  \centering
  \includegraphics[width=\textwidth]{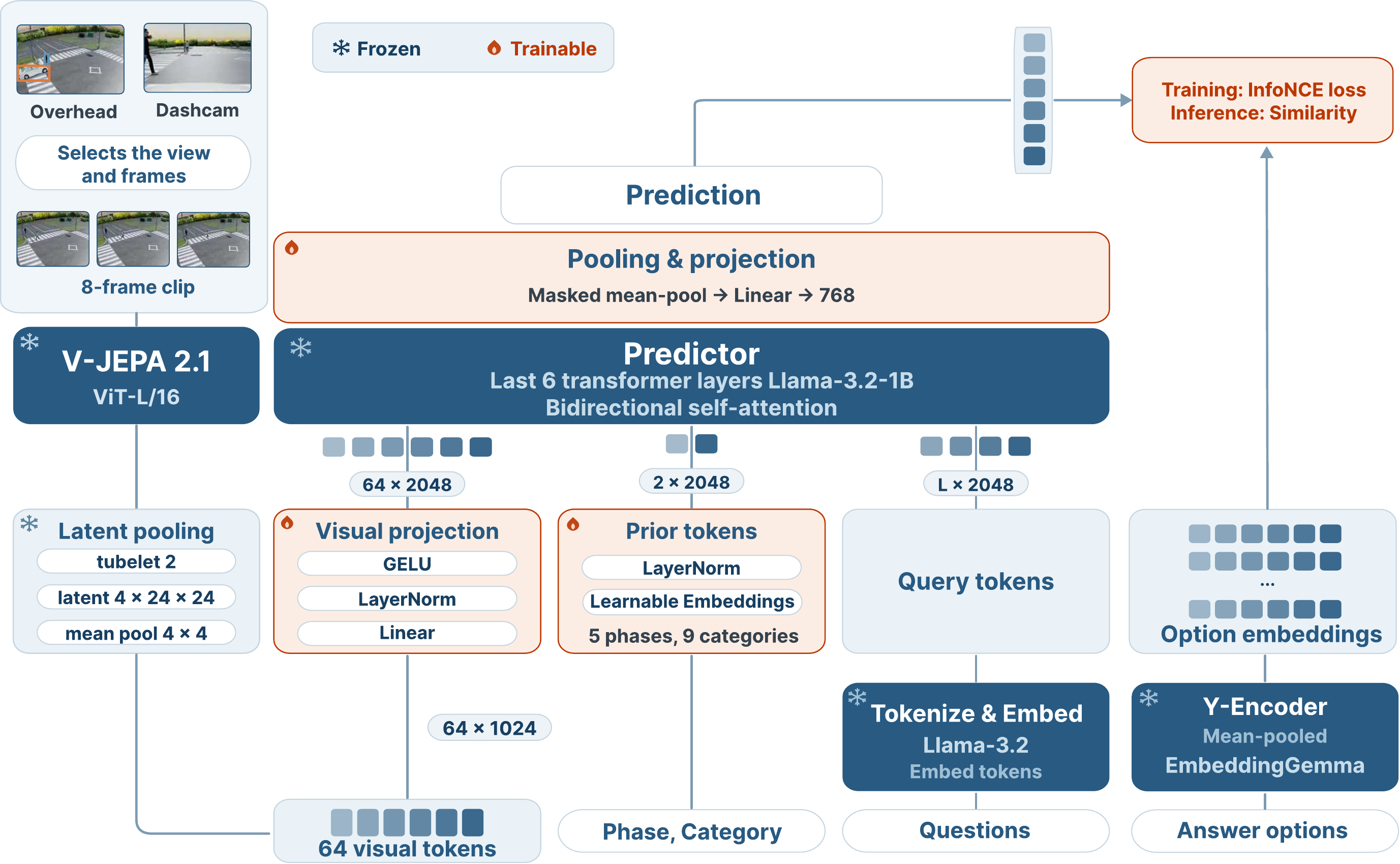}
  \caption{\textbf{Answer prediction.}
    The selected camera video is encoded by a frozen V-JEPA~2.1.
    The video representation is combined with the question and mapped into the
    answer embedding space. Cosine scores provide the InfoNCE objective during
    training and the candidate scores for structured decoding during inference.
    Only the modules shown in orange are optimized on SynWTS.
  }
  \label{fig:model}
\end{figure}

\begin{figure}[t]
  \centering
  \includegraphics[width=\textwidth]{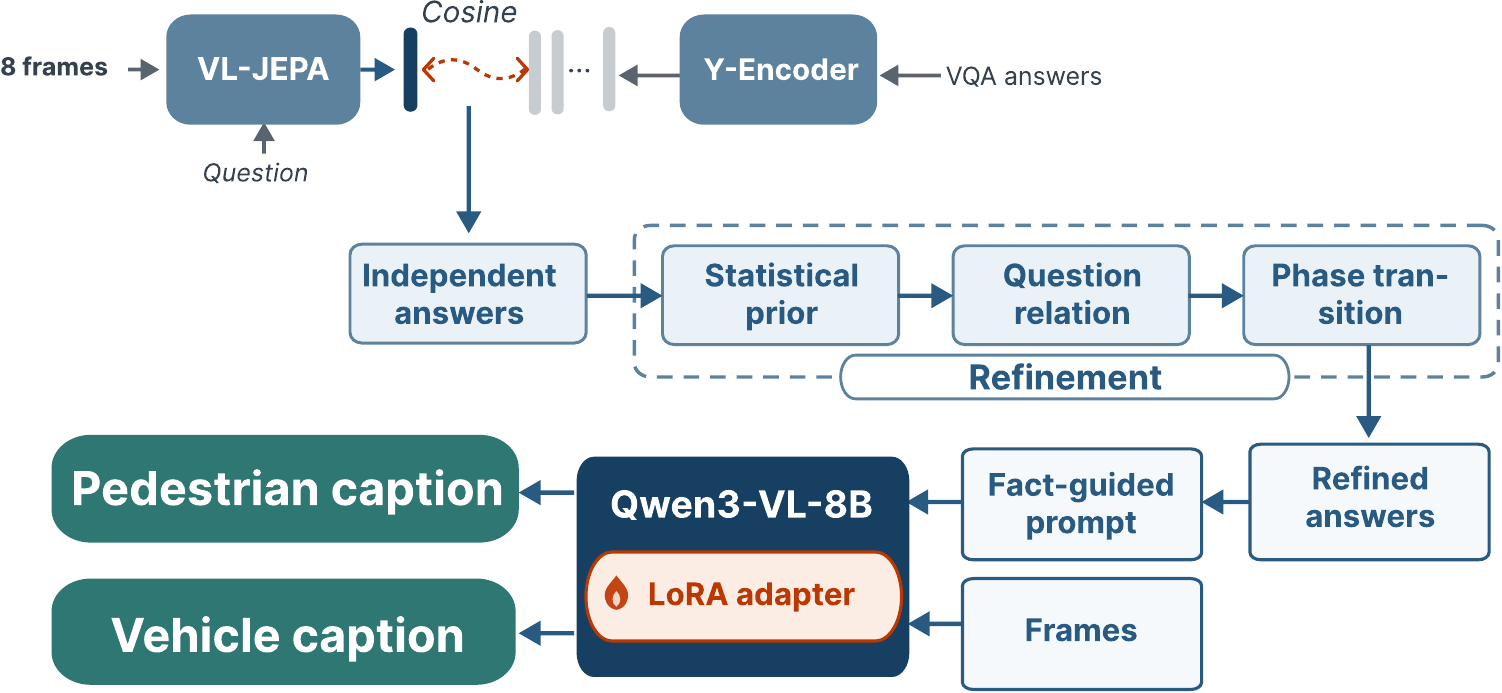}
  \caption{\textbf{Refinement and caption generation.} The independent VQA
    answers pass through the training-free refinement (statistical prior, question
    relation, and phase transition), and the refined answers become facts that,
    together with the frames, guide Qwen3-VL-8B to write the pedestrian and
    vehicle captions.}
  \label{fig:decoder}
\end{figure}

Figure~\ref{fig:model} illustrates the overall pipeline, which consists of five stages. For each question, we first select the camera view that best captures the relevant actors and encode the corresponding event clip using a frozen V-JEPA 2.1 backbone. Inspired by VL-JEPA, the resulting latent representation is conditioned on the question, event phase, and category, allowing the answer to be retrieved from embedded candidate responses rather than generated autoregressively. Since questions are answered independently, the predictions may violate logical relationships or temporal consistency across event phases. To address this issue, we introduce a training-free semantic refinement module that combines statistical priors, directed message passing, and temporal Viterbi decoding~\cite{viterbi1967} to enforce relational and temporal consistency. The refined answers are then normalized into structured facts about the environment, pedestrians, and vehicles, which serve as grounded semantic guidance for Qwen3-VL to generate the final captions~(Fig.~\ref{fig:decoder}).

Given the limited size of the Track 2 dataset, directly fine-tuning large foundation models (V-JEPA 2.1, Llama-3.2-1B~\cite{llama3}, and EmbeddingGemma) is impractical and may lead to overfitting rather than improved traffic understanding. We therefore freeze all pretrained backbones and train only lightweight projection, normalization, and conditioning modules to adapt the information flow between components. This parameter-efficient strategy reduces training costs while retaining the generalizable knowledge acquired during large-scale pretraining.

\subsection{View Selection and Temporal Windowing}
\label{sec:view-selection}

For each question, the corresponding event phase is captured from multiple camera viewpoints, but not all views provide equally informative evidence for semantic understanding. Depending on the viewpoint, the relevant pedestrian or vehicle may appear distant, partially occluded, or provide insufficient visual cues for identifying its state and behavior. To identify the most relevant perspective, we apply a straightforward, annotation-based selection rule. Questions associated with the driver’s viewpoint are answered using the ego camera view. For other questions, we examine the overhead cameras during the queried event phase and select the camera containing the largest bounding box of the relevant pedestrian or vehicle.

After selecting the camera view, we further extract an actor-centric temporal window to remove unrelated frames before encoding. During training, the temporal window is defined by the first and last frames in which the queried actor is annotated with a bounding box within the selected event phase. If this interval is too short, we extend it by four frames at both ends to provide additional temporal context, while remaining within the phase boundaries. During inference, bounding boxes are available only for a small set of sparse key frames. We therefore define the temporal window as spanning from the earliest to the latest annotated frame of the queried actor, with an additional 0.5-second margin on either side. If the resulting window is shorter than one second, we extend it to one second. The window is also constrained to the queried event phase. In both settings, scene-level environment questions are the sole exception, since they are not associated with any specific event phase, and their temporal window therefore spans the entire clip.

Finally, the resulting sequence is then uniformly sampled to eight frames before being processed by V-JEPA, providing a fixed-length input while reducing the computational cost. Bounding box annotations are strictly used for view selection and temporal cropping and are not provided to the model.

\subsection{Encoding with V-JEPA}
\label{sec:temporal-encoding}

Each selected frame is resized to 384×384 without cropping and normalized using ImageNet statistics. We employ a frozen V-JEPA 2.1 ViT-L/16 encoder~\cite{vjepa21} to extract spatiotemporal visual representations. For an eight-frame input clip, the encoder produces four temporal feature maps with a 24×24 spatial resolution, where each spatial location is represented by a 1,024-dimensional feature vector.

The resulting visual representation contains dense spatial tokens that are rich in information but excessive for the subsequent semantic reasoning module. To efficiently adapt the visual features to the downstream task, we apply adaptive average pooling independently to each temporal feature map, reducing the spatial resolution from 24×24 to 4×4 while maintaining the temporal structure. The resulting four temporal grids are flattened into 64 visual tokens, which are then projected from 1,024 to 2,048 dimensions through a trainable linear layer to match the hidden dimension of Llama-3.2-1B. Layer normalization, GELU activation, and dropout are subsequently applied to improve feature adaptation and stability.

\subsection{Question-guided Semantic Reasoning}
\label{sec:answer-prediction}

The benchmark questions are designed to evaluate different aspects of traffic scene understanding through a structured reasoning process. Each question is associated with one of five predefined phases: Pre-recognition, Recognition, Judgment, Action, Avoidance, or related to Environment. We also group the questions into nine semantic categories according to their queried attributes, as listed in Table~\ref{tab:semantic-categories}. The phase and category annotations are used as task-aware priors during question encoding. Lester et al. demonstrated that a similar use of learned conditioning tokens can effectively adapt frozen language models to downstream tasks~\cite{prompttuning}.

\begin{table}[h]
  \centering
  \caption{\textbf{Semantic question categories.} The benchmark questions are
    assigned to nine categories according to the attributes they query.}
  \label{tab:semantic-categories}
  \scriptsize
  \setlength{\tabcolsep}{2.5pt}
  \renewcommand{\arraystretch}{1.08}
  \resizebox{\textwidth}{!}{%
  \begin{tabular}{@{}c l l@{}}
    \toprule
    \textbf{No.} & \textbf{Category} & \textbf{Queried attributes} \\
    \midrule
    1 & Environment
      & Pedestrian appearance and scene, road, traffic, and obstacle attributes \\
    2 & Pedestrian orientation
      & Pedestrian body orientation and direction of travel \\
    3 & Pedestrian position
      & Relative position between the pedestrian and vehicle \\
    4 & Pedestrian to vehicle distance
      & Relative distance between the pedestrian and vehicle \\
    5 & Pedestrian gaze
      & Line of sight, vehicle awareness, and visual status \\
    6 & Pedestrian behavior
      & Coarse and fine-grained pedestrian actions \\
    7 & Pedestrian speed
      & Pedestrian speed \\
    8 & Vehicle orientation
      & Vehicle field of view \\
    9 & Vehicle behavior
      & Vehicle action \\
    \bottomrule
  \end{tabular}
  }
\end{table}

For each question, linguistic tokens are extracted using the frozen Llama-3.2-1B embedding layer and truncated to a maximum length of 64 tokens. Meanwhile, the question phase and category indices into learnable token embeddings, to encode task-specific priors. The visual representation, phase tokens, category tokens and question tokens are subsequently concatenated into a unified sequence and fed into the predictor.

Following the bidirectional predictor design of VL-JEPA, the input sequence is processed by the final six frozen transformer layers of Llama-3.2-1B with the causal mask removed. Since the predictor performs latent representation inference rather than autoregressive generation, causal attention is unnecessary. Removing the mask enables bidirectional interactions among visual tokens, question tokens, phase tokens and category tokens. After the final normalization, all valid output tokens are mean-pooled and mapped through a trainable 2,048-to-768 projection. The resulting vector $\hat{\mathbf{y}}_i \in \mathbb{R}^{768}$ represents the predicted semantic answer embedding conditioned on the video input and question $i$.

By embedding both predictions and candidate answers into a shared semantic space, answer prediction is formulated as an embedding retrieval problem rather than an autoregressive text generation task. For question $i$, let $K_i$ be the number of valid candidate answers and let $o_{i,k}$ denote the text of its $k$-th candidate, where $1 \leq k \leq K_i$. Each candidate is independently encoded using the frozen 300-million-parameter EmbeddingGemma encoder $e(\cdot)$~\cite{embeddinggemma}. The non-padding token representations are mean-pooled to produce $e(o_{i,k}) \in \mathbb{R}^{768}$, with each candidate text truncated to a maximum of 64 tokens. The predicted embedding is compared with every valid candidate embedding using cosine similarity, and the candidate with the highest score is selected:
\begin{equation}
  s_{i,k} = \cos\!\big(\hat{\mathbf{y}}_i,\, e(o_{i,k})\big),
  \qquad
  k_i^{\star} = \operatorname*{arg\,max}_{1 \leq k \leq K_i} s_{i,k},
  \qquad
  a_i = o_{i,k_i^{\star}},
\end{equation}
where $s_{i,k}$ is the cosine similarity score, $k_i^{\star}$ is the index of the highest-scoring candidate (the superscript $\star$ denotes the selected optimum), and $a_i$ is the corresponding textual answer. The comparison depends on the content of each answer rather than its option label or position in the list. Changing the candidate order therefore does not change the semantic comparison. No textual answer is generated at this stage; instead, the similarity scores of all candidates are retained for the subsequent consistency refinement process.

The predictor is trained with an InfoNCE objective~\cite{infonce} to align the predicted representation with the semantic embedding space of candidate answers. Let $k_i^{+}$ denote the index of the ground-truth candidate for question $i$. The loss is defined as
\begin{equation}
  \mathcal{L}_{\mathrm{VQA},i}
  = -\log
  \frac{\exp\!\big(s_{i,k_i^{+}}/\tau\big)}
       {\displaystyle\sum_{k=1}^{K_i} \exp\!\big(s_{i,k}/\tau\big)},
\end{equation}
where $\tau = 0.07$ is the temperature coefficient. The candidate indexed by $k_i^{+}$ is the positive target, while all other candidates for question $i$ serve as negatives. This objective encourages the predicted representation to approach the semantic meaning of the correct answer while separating it from competing interpretations of the same question. By using alternative candidates for the same question as negative samples, the predictor is required to distinguish fine-grained attribute differences rather than only separating unrelated answer concepts.

\subsection{Semantic Consistency Refinement}
\label{sec:structured-decoding}

Although the predictor estimates each question independently, traffic semantics
are inherently structured. Valid answers should not only be supported by visual
evidence, but also be statistically plausible, semantically consistent with
related questions, and temporally coherent across event phases. We therefore
apply a training-free refinement during inference, with all statistics estimated
from SynWTS. It re-ranks only the candidates
the model is unsure about, applying a statistical prior, relational message
passing, and temporal Viterbi decoding in sequence.

\subsubsection{Statistical prior.}
For a given question and event phase, some answers are far more frequent than
others. We adjust each candidate's cosine score by its empirical log probability
under SynWTS, smoothed with Laplace smoothing~\cite{chen1996} so that unseen answers retain a
small nonzero mass. The refined score is
\begin{equation}
  s^{(1)}_i(a) = s_i(a)
  + \gamma \, \log P(a \mid q_i)
  + \gamma_{\varphi} \, \log P(a \mid q_i, \varphi_i),
\end{equation}
where $s_i(a)$ is the cosine similarity score of candidate answer~$a$, $q_i$ and
$\varphi_i$ denote the corresponding question and event phase, and $P(\cdot)$
is the empirical probability estimated from SynWTS. The weights $\gamma=0.02$
and $\gamma_{\varphi}=0.12$ are intentionally small, allowing the priors to
support uncertain predictions without overriding the visual evidence.

\subsubsection{Relational consistency.}
Independent predictions may violate semantic relationships between related
questions, such as distance agreement, inverse relative positions, body
orientation, and movement direction. For example, if the pedestrian is predicted
to be \emph{close} to the vehicle, the vehicle must be equally \emph{close} to
the pedestrian, since both questions measure the same physical gap. As another
example, if the pedestrian is predicted to be \emph{in front of the vehicle}, the
vehicle must be the spatial inverse, \emph{behind the pedestrian}. To address this issue, we perform one
directed message-passing step over a predefined graph of question relations. For
each relation $u\!\to\!v$, the compatibility between candidate answers is
estimated from SynWTS with Lidstone smoothing~\cite{chen1996} as the empirical conditional
probability $C_{uv}(a\mid o_k)$, where $o_k$ is an answer candidate of source
question~$u$ and $a$ is a candidate answer of target question~$v$. When an
observed source answer has no target occurrences, compatibility is assigned a
small probability of $10^{-4}$. Unseen source answers are assigned a uniform
compatibility distribution.

The message from question $u$ to question $v$ is computed in the log domain as
\begin{equation}
  m_{u\to v}(a)
  = \log \sum_{o_k\in\mathcal{A}_u}
    \exp\!\left(\frac{s^{(1)}_u(o_k)}{\tau_m}
    + \log C_{uv}(a\mid o_k)\right),
\end{equation}
where $\mathcal{A}_u$ denotes the candidate set of question~$u$,
$s^{(1)}_u$ is the score after refinement using statistical priors, and $\tau_m=0.32$
controls the message sharpness. The target score is then updated as
\begin{equation}
  s^{(2)}_i(a) = s^{(1)}_i(a) + \alpha w_{uv}m_{u\to v}(a),
\end{equation}
where $\alpha=0.8$ controls the overall contribution of relational information
and $w_{uv}$ assigns a separate weight to each relation type. A single shared
weight would assign too much importance to loose relations and too little
importance to near-deterministic ones,
so each relation is weighted independently, tuned on the SynWTS validation split to match
how reliably it holds in the data. This update increases
the confidence of answers supported by related questions while reducing scores
of semantically inconsistent predictions.

\subsubsection{Temporal consistency.}
Questions describing the same attribute across multiple event phases should
evolve smoothly over time. We therefore jointly decode repeated questions using
Viterbi decoding. Initialized as
$\delta_1(a)=s^{(2)}_1(a)$, the optimal cumulative score for candidate~$a$ at
phase~$t$ is
\begin{equation}
  \delta_t(a) = s^{(2)}_t(a)
  + \max_{b} \Big[\, \delta_{t-1}(b)
    + \beta \, \ell_q\!\big(a \mid b, \varphi_{t-1}, \varphi_t\big)
    + \lambda \, \mathbf{1}[a = b] \,\Big],
\end{equation}
where $\ell_q$ is the transition log probability estimated from SynWTS,
$\varphi_{t-1}$ and $\varphi_t$ denote the previous and current event phases,
$\mathbf{1}[\cdot]$ is the indicator function, $\beta=0.8$ controls the
contribution of transition probabilities, and $\lambda=0.2$ encourages answer
persistence. After the loop completes, the temporal score of question~$i$ is its
cumulative score at that phase,
\begin{equation}
  s^{(3)}_i(a) = \delta_t(a).
\end{equation}
Temporal decoding is applied only to pedestrian behavior, body
orientation, relative position, and gaze, where meaningful temporal evolution
exists. Intuitively, $\ell_q$ favors answer changes that are common between consecutive
phases in SynWTS and penalizes ones that almost never happen, so decoding prefers
sequences that evolve the way attributes usually do. When SynWTS has no examples
for a phase pair, the term stays neutral and lets the model score alone decide.
Decoding is applied only when a question recurs in at least two phases of the
same scenario, since a single-phase chain has no transition to decode.
The optimal answer sequence is recovered by backtracking from the
highest-scoring candidate at the final phase, and the selected candidate is
promoted above the remaining options so that the model score remains the primary
signal.

\subsection{Semantic Guided Caption Generation}
\label{sec:caption-generation}

\begin{figure}[!ht]
  \centering
  \includegraphics[width=.76\textwidth]{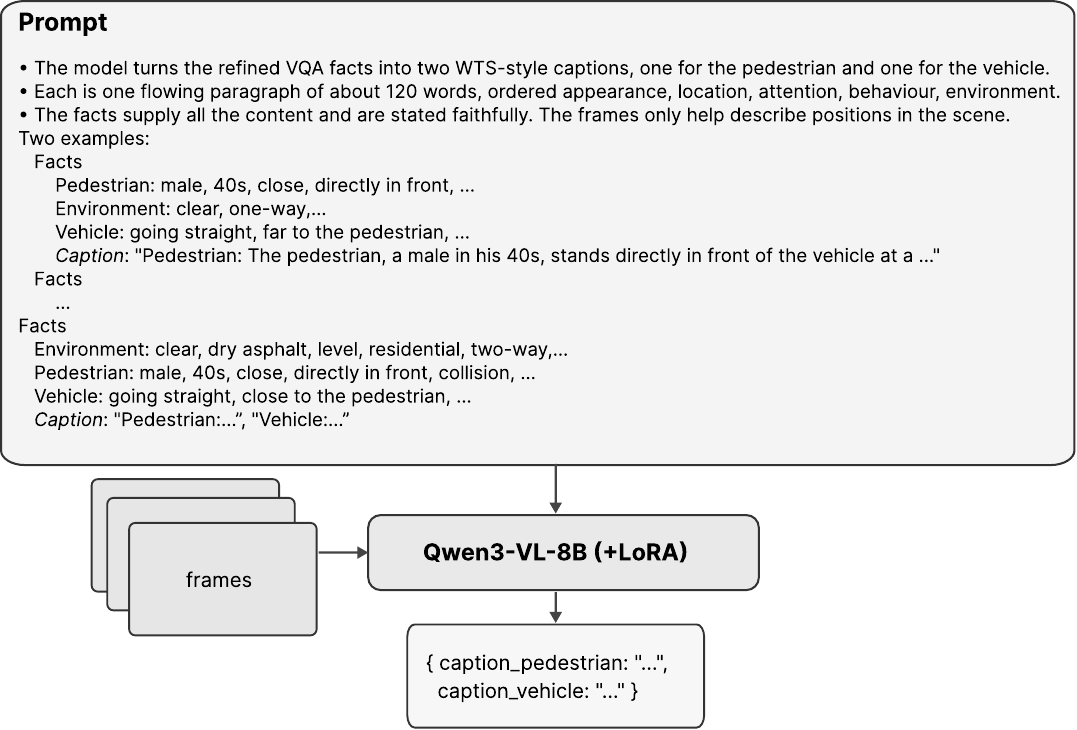}
  \caption{Structure of the caption generation prompt. The refined VQA facts and
  the sampled frames are both fed to Qwen3-VL-8B, which returns the pedestrian
  and vehicle captions as a single JSON object.}
  \label{fig:prompt}
\end{figure}

The refined answers are normalized into structured facts about the environment,
pedestrians, and vehicles. For each event phase, we aggregate the facts inferred
from all associated questions. These facts, together with sampled traffic frames,
are given to Qwen3-VL-8B~\cite{qwen3vl} to generate the caption. This design is supported by
prior work on fact-augmented captioning. Kachhadiya et al.~\cite{kachhadiya2025} likewise
reformulate QA outputs as facts and use these fact-augmented inputs for
structured traffic captioning. From the broader perspective of autonomous driving,
DriveVLM~\cite{drivevlm} organizes VLM reasoning into scene description, scene analysis,
and hierarchical planning, demonstrating that a guided scene description can
capture task-relevant information that supports downstream driving decisions.
Accordingly, our structured facts direct the model toward benchmark-relevant
attributes, while the visual frames supply complementary contextual details.
The pedestrian and vehicle captions are produced together from a single prompt,
whose overall layout is shown in Fig.~\ref{fig:prompt}.

The caption generator is a LoRA-adapted Qwen3-VL-8B~\cite{qwen3vl,lora}. Its inputs are three frames
sampled uniformly across the event phase, the refined VQA facts, and a guided
prompt in which the facts are authoritative and the frames only fill in details the facts
omit, such as scene layout and weather. The LoRA is trained for three epochs on
SynWTS data only, using rank $16$, scale $32$, and dropout $0.05$, with the base
weights kept fixed.

The generated caption is checked against the answer values in the prompt and
regenerated up to four times when required information is missing, after which
deterministic grammar and formatting cleanup is applied. If a segment has no VQA
facts, the sampled overhead and ego-camera frames are provided without the
structured fact blocks as an image-only fallback.

\section{Experiments}
\label{sec:experiments}

\subsection{Datasets}
\label{sec:exp-data}

Track~2 targets the WTS traffic safety benchmark~\cite{wts}, a real dataset of
pedestrian--vehicle interactions recorded simultaneously from several overhead
(CCTV) cameras and a vehicle-mounted camera. Each interaction is split into five
event phases (pre-recognition, recognition, judgement, action, and avoidance) and
annotated per phase with multiple-choice questions on pedestrian, vehicle, and
environment attributes and with reference pedestrian and vehicle captions.
Bounding boxes accompany the videos. The public test set contains $84$ scenarios,
$414$ caption segments, and $4{,}501$ questions, and its labels are withheld for
the official ranking.

To provide labels at scale, the challenge releases SynWTS~\cite{PatilAICity26}, a fully labeled
synthetic digital twin that reproduces the same interactions under the same
annotation schema (questions, captions, phases, camera views, and bounding
boxes)~\cite{aicity2026,Tang26AICity26}. It comprises $249$ scenarios ($167$ for training and $82$ for
validation) with $11{,}733$ questions, each again captured from up to four
overhead views.

\subsection{Evaluation Metrics}
\label{sec:exp-metrics}
The organizers score the two tasks with separate official metrics. VQA is
evaluated by multiple-choice accuracy, and captions by BLEU-4, METEOR, ROUGE-L,
and CIDEr. The final ranking uses a single S2 score, defined as the average of
the VQA result and the caption result, so the two tasks contribute equally.

\subsection{Implementation Details}
\label{sec:exp-impl}

\subsubsection{Training configuration.}
The VQA predictor is trained with AdamW, and the caption adapter with LoRA on
SynWTS frames and facts paired with the reference captions. Table~\ref{tab:implementation} lists the
full training configuration.

\begin{table}[!ht]
  \centering
  \caption{\textbf{Implementation settings.}}
  \label{tab:implementation}
  \scriptsize
  \setlength{\tabcolsep}{3pt}
  \renewcommand{\arraystretch}{1.05}
  \begin{tabular}{@{}lll@{}}
    \toprule
    \textbf{Component} & \textbf{Setting} & \textbf{Value} \\
    \midrule
    \textbf{VQA predictor}
      & Training & AdamW, 16 epochs \\
      & Learning rate / schedule
        & $1.2\times10^{-4}$ / cosine decay / 300 warmup steps \\
      & Batch size & 4 \\
      & Dropout (visual / question) & 0.12 / 0.08 \\
      & InfoNCE temperature & 0.07 \\
    \addlinespace[2pt]
    \textbf{Caption adapter}
      & Training & LoRA, 3 epochs \\
      & Rank / scale / dropout & 16 / 32 / 0.05 \\
      & Learning rate & $10^{-4}$ \\
      & Batch size / maximum length & 16 / 2,048 tokens \\
    \addlinespace[2pt]
    \textbf{Structured decoder}
      & Training & None \\
      & Statistics & Estimated from SynWTS \\
    \bottomrule
  \end{tabular}
\end{table}

\subsubsection{Hardware.}
The VQA predictor is trained and evaluated on a single NVIDIA GeForce RTX~5060~Ti,
while the caption adapter is trained and run on a single NVIDIA RTX~6000.

\subsection{Official Leaderboard Results}
\label{sec:exp-main}

We first compare our system with the other Track~2 teams on the official public
leaderboard. As shown in Table~\ref{tab:main}, it ranks first overall with an S2 of
$60.09$ and a VQA accuracy of $87.09\%$, and scores best on every individual
metric among the top five. Its lead is $2.75$ S2 points over the second team and
is clearest on captions, where our CIDEr of $0.84$ exceeds the next best value
of $0.77$.

This suggests that guiding the caption model with the VQA answers,
rather than letting it describe the frames freely, improves caption quality.
The model still reads the video frames, but the answers steer it toward the
queried attributes. Because the system is trained only on synthetic data,
ranking first on the real test set also confirms that it transfers well across
the sim-to-real gap.

\begin{table}[!ht]
  \centering
  \caption{\textbf{Official public leaderboard.} The five highest-ranked teams
    are shown.}
  \label{tab:main}
  \scriptsize
  \setlength{\tabcolsep}{2.6pt}
  \begin{tabular}{clcccccc}
    \toprule
    Rank & Team & S2 & BLEU-4 & METEOR & ROUGE-L & CIDEr & Acc. (\%) \\
    \midrule
    \textbf{1} & \textbf{Latent Painter - UTE} & \textbf{60.0853} &
      \textbf{0.2798} & \textbf{0.4624} & \textbf{0.4969} &
      \textbf{0.8396} & \textbf{87.0918} \\
    2 & UIT-Kitchen & 57.3307 & 0.2658 & 0.4276 & 0.4595 & 0.5833 & 84.3812 \\
    3 & KZ6         & 56.7949 & 0.2540 & 0.4241 & 0.4471 & 0.7691 & 83.5370 \\
    4 & MobilityAI  & 55.5901 & 0.2532 & 0.4233 & 0.4466 & 0.7601 & 81.2042 \\
    5 & Snow leopard& 55.5768 & 0.2438 & 0.4247 & 0.4446 & 0.7243 & 81.5152 \\
    \bottomrule
  \end{tabular}
\end{table}

\subsection{Effect of the Semantic Consistency Refinement}
\label{sec:exp-decoder}

Because the predictor answers each question independently, its outputs can
contradict one another or change implausibly across phases. This refinement
resolves these inconsistencies at inference time using only SynWTS statistics,
and it raises VQA accuracy from
$81.28\%$ to $87.09\%$ on the official test set, a training-free
gain of $5.81$ points. Table~\ref{tab:decoder} reports the accuracy after each
correction is added, every row scored on the test set by the official evaluation
server. The statistical prior, which only reflects how often
each answer appears in SynWTS, contributes just $+0.36$ points, whereas the
relational and temporal refinement steps do almost all the work, adding
$+1.63$ and $+3.82$ for the cumulative $+5.81$.

The three corrections differ in how well they transfer. The statistical prior only
reflects how frequent each answer is in SynWTS, a dataset-specific signal that is
unlikely to hold on real video, so it is weighted low and adds little. The
relational and temporal steps instead encode the structure of traffic itself,
namely the logical agreement between paired questions such as inverse positions
and distance, and the smooth change of an attribute across phases. These hold
regardless of appearance and account for almost all of the gain, the temporal
step most of it. Because every correction is bounded by the model's own scores
and acts only on low-margin predictions, it fixes genuine errors where the model
is unsure and leaves confident answers untouched, which is why a purely synthetic
refinement still transfers.

\begin{table}[!ht]
  \centering
  \caption{\textbf{Effect of the refinement.} Each row adds one correction on top
    of the rows above. $\Delta$ is the gain over the previous row. All scores come
    from the official evaluation server.}
  \label{tab:decoder}
  \footnotesize
  \setlength{\tabcolsep}{10pt}
  \begin{tabular}{lcc}
    \toprule
    Configuration & Acc.\ (\%) & $\Delta$ \\
    \midrule
    Model alone (no refinement)              & 81.28 & -- \\
    $+$ statistical prior                    & 81.64 & $+0.36$ \\
    \quad $+$ question relation              & 83.27 & $+1.63$ \\
    \quad\quad $+$ phase transition (temporal) & \textbf{87.09} & $\mathbf{+3.82}$ \\
    \bottomrule
  \end{tabular}
\end{table}

\subsection{Caption Configurations}
\label{sec:exp-caption}

Once the refined answers are available, the remaining question is how best to
turn them into captions. We compare two options that share the same grounded
facts and differ only in the language model. The first prompts a frozen
Qwen3-VL-8B with a few SynWTS examples that pair facts with their reference
captions. The second adds a
lightweight LoRA adapter trained on SynWTS frames and facts paired with the
reference captions. To keep the comparison about language quality alone, VQA
accuracy is held fixed at $87.09\%$ for both rows, so any difference in S2 comes
from the captions. As shown in Table~\ref{tab:caption-config}, LoRA improves every
caption metric, most clearly CIDEr from $0.58$ to $0.84$, and lifts S2 from
$57.70$ to $60.09$.

The adapter is trained on synthetic frames but evaluated on real ones, so a gain
measured on the real test set indicates that adaptation did not overfit the
simulated appearance. We attribute this to the prompt, which keeps
appearance-dependent attributes on the facts and leaves the frames only the
scene details that the digital twin reproduces faithfully. The frozen setup
alone already scores competitively, which shows that the refined answers, not
the caption model, carry most of the content.

\begin{table}[!ht]
  \centering
  \caption{\textbf{Effect of LoRA on caption generation performance on the test
    set.}}
  \label{tab:caption-config}
  \small
  \setlength{\tabcolsep}{3.5pt}
  \begin{tabular}{lcccccc}
    \toprule
    Configuration & Acc. (\%) & BLEU-4 & METEOR & ROUGE-L & CIDEr & S2 \\
    \midrule
    Qwen3-VL-8B (base)   & 87.0918 & 0.2221 & 0.4131 & 0.4388 & 0.5846 & 57.7025 \\
    Qwen3-VL-8B $+$ LoRA & 87.0918 & 0.2798 & 0.4624 & 0.4969 & 0.8396 & \textbf{60.0853} \\
    \bottomrule
  \end{tabular}
\end{table}

\section{Conclusion}
\label{sec:conclusion}

We presented a two-stage framework for sim-to-real traffic scene understanding
trained entirely on SynWTS. A frozen V-JEPA 2.1 encoder and a Llama-based
predictor retrieve structured VQA answers. A training-free module refines these
answers using statistical, relational, and temporal consistency. The refined
answers and sampled video frames are then provided to a LoRA-adapted Qwen3-VL-8B
for grounded caption generation.


%
%
\bibliographystyle{splncs04}
\bibliography{main}
\end{document}